\pdfoutput=1

\documentclass[11pt]{article}

\usepackage{coling}
\usepackage{times}
\usepackage{latexsym}
\usepackage{comment}
\usepackage{tcolorbox}
\usepackage{booktabs}
\usepackage{xcolor}
\usepackage{float} 
\usepackage[T1]{fontenc}

\usepackage[utf8]{inputenc}

\usepackage{microtype}

\usepackage{inconsolata}

\usepackage{graphicx}

\title{Charting the Future: Using Chart Question-Answering for Scalable Evaluation of LLM-Driven Data Visualizations}

\author{James Ford, Xingmeng Zhao, Dan Schumacher, \and Anthony Rios \\
  Department of Information Systems and Cyber Security \\
  The University of Texas at San Antonio  \\
  \texttt{\{james.ford, anthony.rios\}@utsa.edu}\\}

\usepackage{subcaption}

\begin{document}
\maketitle
\begin{abstract}
We propose a novel framework that leverages Visual Question Answering (VQA) models to automate the evaluation of LLM-generated data visualizations. Traditional evaluation methods often rely on human judgment, which is costly and unscalable, or focus solely on data accuracy, neglecting the effectiveness of visual communication.  By employing VQA models, we assess data representation quality and the general communicative clarity of charts. Experiments were conducted using two leading VQA benchmark datasets, ChartQA and PlotQA, with visualizations generated by OpenAI's GPT-3.5 Turbo and Meta's Llama 3.1 70B-Instruct models. Our results indicate that LLM-generated charts do not match the accuracy of the original non-LLM-generated charts based on VQA performance measures. Moreover, while our results demonstrate that few-shot prompting significantly boosts the accuracy of chart generation, considerable progress remains to be made before LLMs can fully match the precision of human-generated graphs. This underscores the importance of our work, which expedites the research process by enabling rapid iteration without the need for human annotation, thus accelerating advancements in this field.

\end{abstract}

\section{Introduction}

Data analytics is integral to modern organizations, enabling informed decision-making by interpreting complex datasets. Effective data visualization transforms vast amounts of information into actionable insights, but the increasing volume and complexity of data can overwhelm organizational staff. Many individuals lack the technical skills needed to generate meaningful visualizations, creating a barrier between data availability and utilization.

Recent advancements have seen Large Language Models (LLMs) applied to data visualization tasks, allowing users to create visual representations through natural language queries~\cite{masry-etal-2022-chartqa,Methani_2020_WACV}. While this progress is promising, evaluating the quality of LLM-generated visualizations poses significant challenges. Traditional evaluation methods rely heavily on human judgment to assess data accuracy and the effectiveness of visual communication. This dependence on human evaluators is not only costly but also impractical for scaling across large datasets or diverse visualization types. Subjectivity and variability in human assessments further complicate consistent benchmarking and continuous system development.

An alternative evaluation approach involves regenerating the input data from the visualization and comparing it to the original dataset. Although this method checks for data representation accuracy, it overlooks critical factors such as visual design clarity, interpretability, and the visualization's ability to highlight key insights~\cite{tenney-etal-2020-language}. For example, a chart might accurately reflect the underlying data but fail to convey the intended message due to poor color schemes, misleading labels, or cluttered layouts. Such issues would remain undetected through data regeneration methods, as they focus solely on data fidelity rather than the effectiveness of information communication.

\begin{figure}[t]
\centering
\includegraphics[width=.8\linewidth]{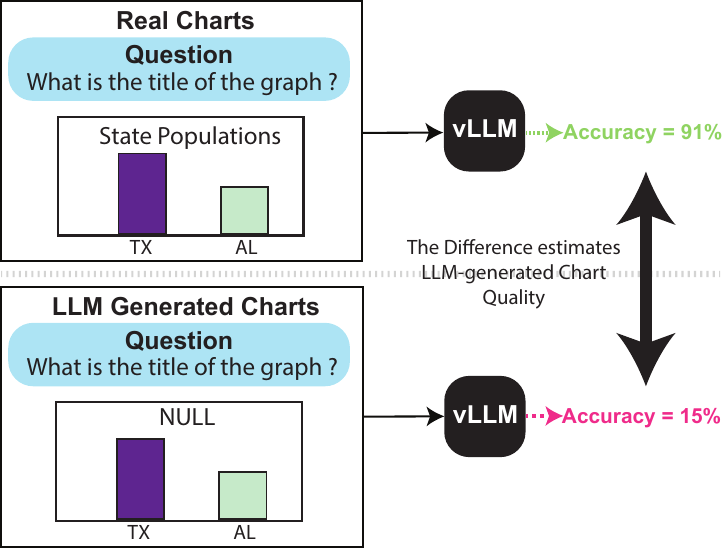}
\caption{An overview of the VQA evaluation process for generated visualizations. The visual LLMs (vLLMs) represent trained models for chat QA. }\vspace{-1em}
\label{fig:intro} \end{figure}

To help address these challenges in evaluating visualizations, we present Visual Question Answering (VQA) as a more comprehensive solution. VQA models can assess both the data and design aspects of a chart by answering questions about its content, offering a deeper evaluation of the visualization’s effectiveness~\cite{liu-etal-2023-matcha, masry-etal-2023-unichart}. For example, a VQA model might respond to queries like ``What trend is depicted in this chart?,'' ``Which category has the highest value?,'' or ``What is the second largest bar?'' These questions assess user interpretation and the visualization's communicative success, going beyond mere data verification. If we can accurately predict the answers to these and similar questions, then the chart should be adequate. This approach aligns closely with the ultimate goal of visualizations: to effectively communicate insights.

Moreover, VQA enables automated evaluation at scale, reducing reliance on human evaluators and facilitating large-scale assessments across various visualization types. VQA provides an evaluation method that mirrors real-world user interaction with visual content by focusing on interpretability and design aspects. An illustration of the VQA process is shown in Figure~\ref{fig:intro} . Specifically, In Figure~\ref{fig:intro}, we provide an example of how our VQA-based evaluation framework distinguishes between real and LLM-generated visualizations. The VQA model answers a question about the title of a graph depicting state populations for Texas and Alabama. In the case of a real chart, the model is able to correctly identify the title, ``State Populations,'' with 91\% accuracy. However, when evaluating an LLM-generated chart, which lacks a title (denoted as ``NULL'') because of issues with the chart generation process, the accuracy drops to 15\%. The performance difference serves as an indicator of the LLM-generated chart's quality. This process highlights how our approach can effectively measure not only data accuracy but also the communicative clarity of LLM-generated visualizations. Overall, with our system, users could develop new chart generation methods and then rapidly iterate them using the VQA accuracy results we propose.

In this paper, we propose a novel framework that leverages chart-based VQA models to automate the evaluation of LLM-generated visualizations. Our contributions are threefold:

\begin{enumerate}
    \item  We develop an automated framework capable of scaling the evaluation process for chart generation, enabling efficient benchmarking across multiple models and datasets.
    \item We demonstrate that VQA models offer unique, context-sensitive feedback compared to data regeneration methods, providing a holistic assessment of both data accuracy and visual communication effectiveness.
    \item We present empirical results that showcase the effectiveness of our approach in large-scale evaluations, comparing multiple LLMs using different prompting strategies.
\end{enumerate}

\section{Related Work}

Data visualization is crucial for interpreting complex information, yet effective visualizations often requires technical expertise~\cite{10.2312/eurovisshort.20171133,dibia-2023-lida}. Text-to-Viz systems (T2V) aim to simplify this process by enabling users to generate visualizations through natural language queries~\cite{9699035,10530359}. This approach addresses challenges non-technical users face, such as steep learning curves and difficulty selecting appropriate visualization methods~\cite{app13127025}. Complete T2V typically consists of two components: data querying and visualization~\cite{affolter_comparative_2019,10530359}. While our focus is on the visualization aspect, the process involves multiple steps, including parsing the input query, identifying data attributes, and choosing appropriate visualization styles~\cite{10121440}. Commercial tools like Tableau and Microsoft Power BI have incorporated limited T2V capabilities~\cite{10121440}.

Challenges for T2V include ambiguity and under-specification in natural language, and users often overestimate the system's capabilities~\cite{8986918}. Early T2V relied on rule-based or template-based methods with shallow parsing techniques~\cite{9699035,10530359}, which struggled with complex data and offered limited efficacy~\cite{hong2023conversationalaithreadsvisualizing}. Advances in Natural Language Processing and deep learning introduced more robust models, such as sequence-to-sequence architectures and pre-trained language models like BERT and T5, improving complexity and robustness but requiring extensive training data~\cite{10530359,10443572,voigt-etal-2023-vist5}. For example, FLAN-T5 has been used to create Vega-Lite specifications for data visualizations~\cite{voigt-etal-2023-vist5,10443572}. Toolkits like NL4DV facilitated the creation of Vega-Lite specifications from natural language~\cite{9222342}. Interactive systems combining text and speech inputs have also been developed~\cite{10.1145/3313831.3376782,9303381}.

LLMs have led to new approaches in generating data visualizations~\cite{dibia-2023-lida,10121440}. LLMs can produce visualization code or images directly from user queries without extensive pre-training~\cite{hong2023conversationalaithreadsvisualizing}. Research has shown that LLMs can match the performance of human data analysts~\cite{cheng-etal-2023-gpt}, though challenges remain~\cite{zhang2024benchmarkingdatascienceagents,gu2024analystsunderstandverifyaiassisted}. OpenAI's Codex has been used to generate visualization interfaces~\cite{chen2022nl2interfaceinteractivevisualizationinterface}, and GPT models have been employed to create Python scripts~\cite{10121440}. Techniques like prompt engineering and chain-of-thought processes have enhanced LLM-generated visualizations~\cite{li2024visualizationgenerationlargelanguage,podo2024vrecslowcostllm4visrecommender}.

Evaluating LLM-generated visualizations is challenging. Traditional methods include comparing generated code to ground-truth specifications, comparing visual outputs, and user satisfaction ratings~\cite{10530359}. However, these often rely on human judgment, which is subjective, costly, and hard to scale~\cite{10121440,10458347,9118800}. Automated efforts primarily focus on checking the syntax of generated code~\cite{8744242}, but code similarity may not be a good indicator of visualization quality~\cite{10670425}. Some frameworks attempt to automate parts of the evaluation but still require human involvement~\cite{10344146,podo2024vievallmconceptualstack,10670425}. Moreover, LLM-generated visualizations can be prone to hallucinations and inconsistencies~\cite{podo2024vievallmconceptualstack,10443572,10121440}, and may depend heavily on user interventions for quality control~\cite{ijgi12070284}. Research suggests that human feedback is preferred over LLM feedback in evaluation~\cite{kim2024goodchatgptgivingadvice}.

VQA on charts offers a promising alternative for automated evaluation chart generation quality instead of using human evaluation only. VQA models answer questions about data and design elements, providing a richer assessment of a chart's effectiveness~\cite{liu-etal-2023-matcha,masry-etal-2023-unichart}. Two main approaches are used: converting charts into underlying tables for text analysis~\cite{han2023chartllamamultimodalllmchart}, and using multimodal models to derive answers directly from the chart. Datasets like ChartQA~\cite{masry-etal-2022-chartqa} and PlotQA~\cite{Methani_2020_WACV} have been instrumental in advancing VQA research. Models such as Unichart~\cite{masry-etal-2023-unichart} and MatCha~\cite{liu-etal-2023-matcha} have achieved high performance on these datasets.

Building on these advancements, our work leverages chart-based VQA (in contrast with general image VQA) models to automate the evaluation of LLM-generated visualizations. By focusing on both data accuracy and visual communication effectiveness, we address the limitations of previous evaluation methods and enable scalable, automated assessment without human intervention.

\section{Data}

\begin{table}[t]
  \centering
  \resizebox{.8\linewidth}{!}{
  \begin{tabular}{lrrr}
    \toprule
    \textbf{Dataset} & \textbf{Charts} & \textbf{Question-Answer Pairs} \\
    \midrule
    ChartQA     & 734    & 1,113     \\
    PlotQA     & 400     & 14,427           \\\bottomrule
  \end{tabular}}
  \caption{Dataset sample statistics.}\vspace{-1em}
  \label{tab:stats}
\end{table}

Our study uses subsets from two datasets: PlotQA~\cite{Methani_2020_WACV} and ChartQA~\cite{masry-etal-2022-chartqa}. The statistics for the subsets of both datasets used in our study are shown in Table~\ref{tab:stats}.

\vspace{2mm} \noindent \textbf{PlotQA.} The PlotQA dataset~\cite{Methani_2020_WACV} is a large-scale benchmark for visual question answering (VQA) over scientific plots, comprising over 224,000 plots and approximately 28.9 million question-answer pairs derived from real-world data sources. It challenges models to interpret complex data visualizations by categorizing questions into three main types: \textit{Structural Understanding}, which involve questions about the overall structure of the plot without requiring quantitative reasoning (e.g., the presence of grid lines or legend labels); \textit{Data Retrieval}, which require extracting specific data values directly from the plot (e.g., reading exact numerical values or labels); and \textit{Reasoning}, which involve numerical reasoning over multiple plot elements or comparative analysis (e.g., performing arithmetic calculations, comparisons, or interpreting trends). This diversity of question types necessitates models that can handle complex reasoning tasks, precise data extraction, and an understanding of structural nuances in various plot types, making PlotQA a challenging and comprehensive dataset for evaluating the reasoning capabilities of models in the context of data visualizations. We sample 14,427 QA pairs from 400 charts because of commercial API costs for chart generation and computational expenses.

\begin{figure*}[t]
    \centering
    \includegraphics[width=1.\linewidth]{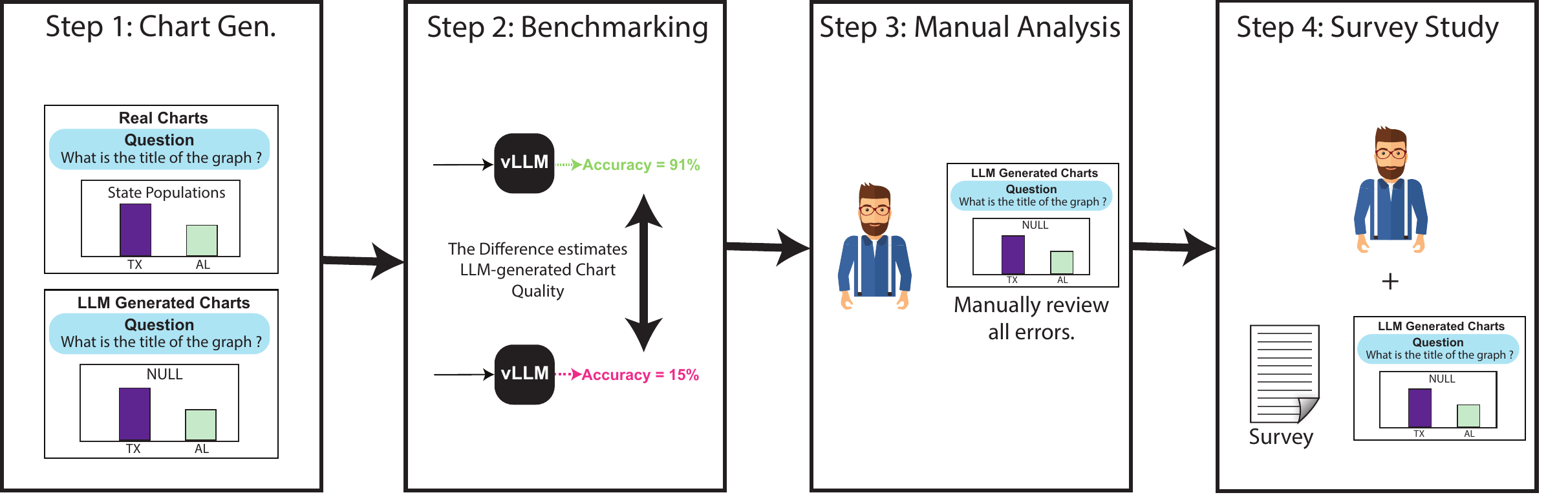}
    \caption{Overall framework for our study, where we perform automatic chart generation, benchmarking using chart question answering, manually analyze errors, and perform a survey on chart quality.}
    \label{fig:overview}
    \vspace{-1em}
\end{figure*}

\vspace{2mm} \noindent \textbf{ChartQA.} ChartQA is a comprehensive benchmark dataset designed to advance research in question answering over data visualizations, specifically focusing on charts like bar graphs, line graphs, and pie charts. It comprises 4,804 charts with 9,608 human-authored question-answer pairs (ChartQA-H) and 17,141 charts with 23,111 machine-generated question-answer pairs (ChartQA-M), resulting in a total of 21,945 charts and 32,719 questions. The charts are collected from diverse real-world sources such as Statista, Pew Research, Our World in Data, and the OECD, ensuring various chart styles, topics, and data representations. ChartQA emphasizes complex reasoning tasks that require models to perform multiple logical and arithmetic operations and to handle open-vocabulary answers derived from chart data, rather than selecting from a fixed set. Many questions also involve visual reasoning, referring to specific visual attributes like color, size, or position of chart elements. Because of the cost of testing commercial models, we sample 734 charts to have around 1,100 question-answer pairs.

\section{Methodology}

We provide a high-level overview of our paper in Figure~\ref{fig:overview}. Overall, our framework has four main components. First, we generate charts using two popular LLMs for both datasets used in our study. Second, we benchmark the chart generation quality of the LLMs using the chart question-answering task. Third, we manually review all contrasting errors (i.e., errors made by one model, but not the other) to ensure the question-answering task is working as expected. Fourth, we perform a small survey study where we have participants review charts and then we compare the results with the benchmarking from Step 2. Steps 1 and 2 are the evaluation framework we are proposing. Steps 3 and 4 are how we evaluate whether the VQA results accurately measure chart quality.

\vspace{2mm} \noindent \textbf{Step 1: Baseline Methods and Dataset Preparation.}

In this step, we use two LLMs to generate data visualizations from a prompt. Specifically, we use OpenAI's GPT-3.5 Turbo-0125 and NeuralMagic's Meta LLaMA 3.1 70B-Instruct-FP8 models~\cite{neuralmagic_MetaLlama3_70B_Instruct_FP8_2024} to produce Python \texttt{matplotlib} code that creates charts based on provided datasets. Our focus is on designing effective prompts for these LLMs using zero-shot and few-shot prompting strategies.

In the zero-shot setting, the LLMs receive only the task instructions and the data without any examples. The system prompt instructs the model to generate Python code for data visualization:
\begin{center}
\vspace{-.25em}
\tcbset{
    colframe=black,
    colback=white,
    boxrule=0.5mm,
    arc=3mm,
    width=1\linewidth,
    boxsep=5pt,
    left=5pt,
    right=5pt,
    top=5pt,
    bottom=5pt
}
\begin{tcolorbox}[colback=blue!5!white,colframe=blue!75!black,title=System Instructions]
\textit{You are a data analyst tasked with creating data visualization plots based on the provided data. Output should be formatted as Python \texttt{matplotlib} code and must include both \texttt{fig.clf()} and the \texttt{bbox\_inches='tight'} parameter. Use the specified title, chart type, and data for the axis labels and counts. Do not use \texttt{list(range)}. Ensure data value labels are on the chart, place legends outside the chart, and save the chart as a PNG file using the specified filename.}
\end{tcolorbox}
\vspace{-.25em}
\end{center}

The data prompt then provides the specific chart details
\begin{center}
\vspace{-.25em}
\tcbset{
    colframe=black,
    colback=white,
    boxrule=0.5mm,
    arc=3mm,
    width=1\linewidth,
    boxsep=5pt,
    left=5pt,
    right=5pt,
    top=5pt,
    bottom=5pt
}
\begin{tcolorbox}[colback=blue!5!white,colframe=blue!75!black,title=Input Query]
    \textit{\texttt{data\_pass} = ``Title: \texttt{title\_text} / Data: \texttt{final\_string} / Chart type: \texttt{figure\_type} / File Name: \texttt{png\_file}''}
\end{tcolorbox}
\vspace{-.25em}
\end{center}
where \texttt{title\_text} is the chart title, \texttt{final\_string} contains the data in text format, \texttt{figure\_type} specifies the chart type (e.g., bar, line), and  \texttt{png\_file} is the desired output file name.
This prompt directs the LLM to generate the appropriate Python code to create the chart as per the given specifications. In the few-shot setting, the system prompt is mentioned once, and the data prompt (Input Query) is repeated for each of the in-context examples.

\vspace{2mm} \noindent \textbf{Step 2: Benchmarking Chart Quality using Question-Answering.}

We used two VQA models to perform the question-answering task. \citet{masry-etal-2023-unichart} developed UniChart, a model with two modules (a chart encoder based on the document image Donut model~\cite{kim2022ocr}, plus a text decoder based on a BART model~\cite{lewis2019bart}) pretrained on more than 600,000 charts, optimized explicitly for the ChartQA dataset. UniChart processes a plot image by first encoding its textual elements (such as legends and axis labels), visual elements (like bars and lines), and the overall layout. It then decodes this information to generate answers based on the content of the plot.
\citet{liu-etal-2023-matcha} created MatCha, adding chart derendering and mathematical reasoning pretraining to the Pix2Struct vision-language model~\cite{lee2023pix2struct}.  Unichart and Matcha were chosen because they both reported high performance on VQA with the ChartQA and PlotQA datasets, respectively.

The VQA models are applied to the charts generated in Step 1 in order to derive listings of generated answers which were then assessed against the ground truth answers from the two datasets. Accuracy is defined as the answer from the VQA models matching the ground truth in the dataset. For text responses, strict accuracy was employed but relaxed accuracy~\cite{masry-etal-2022-chartqa} was used for mathematical and numerical 
responses. Relaxed accuracy is operationalized as accepting numerical values that are within plus or 
minus five percent.

\vspace{2mm} \noindent \textbf{Step 3: Manual Quality Analysis.} We manually reviewed the errors made exclusively by GPT-3.5 Turbo and LLaMA 3.1 to ensure that our VQA evaluation accurately reflects the quality of the charts generated by these models. Using a qualitative coding framework, we categorized these errors to determine whether discrepancies were due to issues in the charts or unrelated factors. This is important because if the VQA errors aren't caused by issues with the chart quality, then using the VQA task to assess the quality of the charts wouldn't be a valid evaluation method.

In our analysis, we identify instances where the VQA model marked answers incorrect for charts from one LLM but correct for the other. We examined each case, categorizing errors as either visualization errors—such as incorrect data representation, mislabeled axes, missing titles, or overlapping labels—or errors due to the VQA model itself, like misinterpretation or ambiguous questions. Because VQA models are not perfect, we expect those issues to impact both models similarly.

By categorizing the errors, we assess whether the VQA task effectively measures chart quality. If most errors stem from visualization issues in the charts, it confirms that VQA is a valid assessment tool. For example, if GPT-3.5 Turbo's chart had poorly presented labels that led the VQA model to misinterpret data, while LLaMA's chart didn’t, the issue would stem from chart quality. This shows that the VQA task reflects chart performance, while also helping us identify the types of errors our framework detects.

\vspace{2mm} \noindent \textbf{Step 4: Survey.}
In this paper, we focus on assessing the quality of charts generated by large language models (LLMs) when converting structured data into visualizations. To evaluate these charts, we designed a set of questions for human participants, aligning them with the question types used in the PlotQA dataset~\cite{Methani_2020_WACV}. The PlotQA dataset categorizes questions into three main types: \textit{Structural Understanding}, \textit{Data Retrieval}, and \textit{Reasoning}. We aim to draw correlations between human assessments and automated evaluation metrics by mapping our human evaluation questions to these types.

The following questions guide our human evaluation, where participants assess the accuracy, readability, and overall usefulness of these charts:

\vspace{2mm}
\noindent \textbf{\textit{Q1: The LLM-generated chart accurately displays a title reflecting the data depicted in the original data file.}}\textbf{(Structural Understanding)} This question corresponds to the \textit{Structural Understanding} type in PlotQA. The title of a chart serves as a crucial summary of the data it represents. This question evaluates whether the LLM can produce a chart title that accurately reflects the underlying data, ensuring that viewers can quickly grasp the content of the visualization.

\vspace{2mm}
\noindent \textbf{\textit{Q2: The X-axis labels on the LLM-generated chart accurately display the labels depicted in the original data file.}}\textbf{(Structural Understanding)} Aligned with \textit{Structural Understanding} in PlotQA, the X-axis often represents categories or time intervals in visualizations. This question focuses on the LLM's ability to correctly generate X-axis labels that are faithful to the labels present in the data file, ensuring correct interpretation of the chart's horizontal dimension.

\begin{table*}
  \centering
  \resizebox{.8\textwidth}{!}{
  \begin{tabular}{lrrrrr}
    \toprule
    \textbf{Model} & \textbf{Overall} & \textbf{Human (n=390)} & \textbf{Augmented (n=723)} &  \textbf{Sim. Score }& \textbf{Exact Match} \\
    \midrule
    Original Charts     & 64.4\    & 35.4\    & 80.1 & 97.8 & 70.6     \\ \midrule
    ChatGPT Zero-Shot  & 42.3\    & 21.5\    & 53.5\     & 97.0 & 30.8         \\
    Llama 3.1 Zero-Shot     & 51.1\    & 34.0\     & 60.9\   & 95.3 & 33.1  \\
    ChatGPT Few-Shot     & 44.2\    & 23.7\     & 55.2\  & 97.2 & 43.5   \\
    Llama 3.1 Few-Shot     & 58.0\    & 37.1\     & 69.9\  & 93.1 & 35.8    \\
    \bottomrule
  \end{tabular}}
  \caption{VQA Results from ChartQA Dataset. The UniChart VQA model was used for these experiments.}
  \label{tab:chartqa}
\end{table*}

\vspace{2mm}
\noindent \textbf{\textit{Q3: The Y-axis labels on the LLM-generated chart accurately display the labels depicted in the original data file.}} \textbf{(Structural Understanding)} Also a \textit{Structural Understanding} question type, the Y-axis typically corresponds to numerical values or other measurements. This question evaluates whether the Y-axis labels in the LLM-generated chart match the values in the data file. Accurate Y-axis labels are essential for interpreting data magnitude and making comparisons across categories.

\vspace{2mm}
\noindent \textbf{\textit{Q4: The data points on the LLM-generated chart accurately display the values depicted in the original CSV data file.}} \textbf{(Data Retrieval)} This question maps to the \textit{Data Retrieval} type in PlotQA. Data point accuracy is critical to ensure the visualization faithfully represents the dataset. It measures whether the individual data points (e.g., bars, lines, or dots) on the chart match the corresponding values in the CSV file, maintaining the integrity of the visualized data.

\vspace{2mm}
\noindent \textbf{\textit{Q5: The LLM-generated chart is easy to read and understand.}} \textbf{(Reasoning)} This question relates to the \textit{Reasoning} type in PlotQA, specifically concerning the user's ability to perform reasoning tasks using the chart. Even if a chart is accurate, it must also be easy for users to interpret. This question evaluates the chart's readability, including clarity of labels, appropriate scaling, and overall visual design. Readability ensures that users can extract insights and perform complex reasoning without unnecessary cognitive effort.

\vspace{2mm}
\noindent \textbf{\textit{Q6: Overall, the LLM-generated chart serves its intended purpose in a satisfactory manner.}} \textbf{(Comprehensive Assessment)} This question encompasses all three PlotQA question types—\textit{Structural Understanding}, \textit{Data Retrieval}, and \textit{Reasoning}. It assesses the overall effectiveness of the chart in fulfilling its intended purpose, whether that be conveying specific trends, making comparisons, or summarizing key data points. Participants evaluate whether the chart helps them interpret and draw meaningful conclusions from the data.

Each of these questions is measured using a 5-point Likert scale, where participants rate their level of agreement with each statement (1 = Strongly Disagree, 5 = Strongly Agree). By mapping our evaluation questions to the PlotQA question types, we aim to correlate human ratings with automated evaluation metrics, providing a comprehensive understanding of how well LLM-generated charts perform in practical scenarios.

\begin{table*}
  \centering
  \resizebox{\textwidth}{!}{
  \begin{tabular}{lrrrrrrrrr}
    \toprule
    \textbf{} & \textbf{} & \textbf{Arithmetic} & \textbf{Comparison} & \textbf{Compound} & \textbf{Data-Retrieval} & \textbf{Min-Max} & \textbf{Structural} \\    
    \textbf{Model} & \textbf{Overall} & \textbf{(n=1,789)} & \textbf{(n=1,044)} & \textbf{(n=2,311)} & \textbf{(n=3,352)} & \textbf{(n=1,146)} & \textbf{(n=4,485)} & \textbf{Sim. Score} & \textbf{Exact Match} \\
    \midrule
     Original Charts    & 80.6\    & 43.6\    & 73.2\  & 71.2\   & 88.3\    & 94.1\  & 92.0\   & 92.2 & 16.0    \\ \midrule
    ChatGPT Zero-Shot   & 60.5\    & 22.7\    & 61.3\  & 54.3\   & 59.6\    & 76.8\  & 74.2\     & 92.5 & 49.3   \\
    Llama 3.1 Zero-Shot   & 54.6\    & 16.9\    & 51.5\  & 49.9\   & 57.2\    & 65.9\  & 67.0\      & 82.7 & 25.0  \\
    ChatGPT Few-Shot   & 62.4\    & 23.8\    & 63.9\  & 54.6\   & 61.9\    & 80.7\  & 75.9\    & 92.3 & 37.0    \\
    Llama 3.1 Few-Shot   & 61.9\    & 23.9\    & 63.8\  & 54.2\   & 62.4\    & 79.7\  & 74.6\  & 93.1 & 35.8      \\
    \bottomrule
  \end{tabular}}
  \caption{VQA Results from PlotQA Dataset. The MatCha VQA model was used for these experiments.}\vspace{-1em}
  \label{tab:plotqa}
\end{table*}

\section{Results}

\vspace{2mm} \noindent \textbf{Implementation Details.} We used the NeuralMagic's Meta LLaMA 3.1 70B-Instruct-FP8 models~\cite{neuralmagic_MetaLlama3_70B_Instruct_FP8_2024} for our prompting-based experiments, running it on two NVIDIA A6000 GPUs. To generate outputs, we set the sampling parameters with a temperature of 0.1, top-p of 0.9, and a maximum token limit of 2,000 to balance coherence and diversity in the model's responses. All experiments were conducted using PyTorch~\cite{paszke2019pytorch} and the Hugging Face Transformers library~\cite{wolf2019huggingface}. We only used three few-shot examples in our experiments.

\begin{table}[t]
  \centering
  \resizebox{\linewidth}{!}{
\begin{tabular}{lrr}
\toprule
\textbf{Error Type} & \textbf{GPT3.5 Errors} & \textbf{Llama Errors} \\
\midrule
VQA Model Error & 38 (22.89\%) & 6 (15.79\%) \\
Actually Correct & 2 (1.20\%) & 1 (2.63\%) \\ \midrule
Bar Boundaries & 2 (1.20\%) & 0 (0.00\%) \\
Category Ambiguous & 10 (6.02\%) & 0 (0.00\%) \\
Colors Not Matching & 6 (3.61\%) & 1 (2.63\%) \\
Dates Errors & 49 (29.52\%) & 11 (28.95\%) \\
Labels Overlapping & 55 (33.13\%) & 12 (31.58\%) \\
Wrong Type of Bars & 3 (1.81\%) & 7 (18.42\%) \\
Chart Not Displaying & 1 (0.60\%) & 0 (0.00\%) \\
\bottomrule
\end{tabular}}
  \caption{Visualization Error Mismatches Between ChatGPT and Llama}\vspace{-1em}
  \label{tab:errors}
\end{table}

\vspace{2mm}
\noindent \textbf{Comparison Metrics.} We also measure the accuracy of the charts by extracting the data depicted in them.  Two metrics were used to gauge the accuracy of the extracted data, comparing the extractions with the ground truth data files from the datasets. \textit{Similarity Score} uses the normalized distance between the extracted values from each LLM-generated chart and the ground truth data files.  This is calculated as the absolute difference between $x_1$ and $x_2$, divided by ($x_1$ + $1e^{-15}$), where $x_1$ is the ground truth data and $x_2$ is the extracted data. The second method, \textit{Exact Match}, computes the percentage of charts whose extracted data tables are exact matches with the ground truth tables. The Python pandas function ``exact'' was applied to each column to verify that the columns were identical (which checks that items match exactly and are in the correct order) and is thus a more challenging metric than the first method.

\begin{table*}
  \centering
    \resizebox{.7\linewidth}{!}{
  \begin{tabular}{lrr}
    \toprule
    \textbf{Dataset / Model} & \textbf{ChatGPT} & \textbf{Llama 3.1} \\
    \midrule
    The LLM-generated chart accurately displays a title reflecting     & 4.82    & 4.88     \\   
       the data depicted in the original CSV data file.          \\ 
    The X-axis labels on the LLM-generated chart accurately displays     & 4.28    & 4.15     \\  
       the labels depicted in the original CSV data file.      \\    
    The Y-axis labels on the LLM-generated chart accurately displays     & 4.80    & 4.91     \\       
       the labels depicted in the original CSV data file.     \\    
    The data points on the LLM-generated chart accurately displays     & 4.85    & 4.75     \\      
       the values depicted in the original CSV data file.      \\  
    The LLM-generated chart is easy to read and understand.     & 3.79    & 3.95     \\      
    The LLM-generated chart is visually appealing.     & 3.68    & 3.68     \\      
    Overall, the LLM-generated chart serves its intended purpose     & 3.85    & 3.87     \\     
       in a satisfactory manner.         \\
    \bottomrule
  \end{tabular}}
  \caption{Human Evaluation Results:  ChartQA Few-Shot}
  \label{tab:human}\vspace{-1em}
\end{table*}

\vspace{2mm} \noindent \textbf{Benchmark Results.} We report the ChartQA benchmark results in Table~\ref{tab:chartqa}. Overall, we make four major findings. First, all performance metrics are lower on the generated charts than on the original charts. At least at a superficial level, this is useful to know that LLMs struggle to generate human-quality charts. The best performance overall performance was by the LLama 3.1 70B model, which achieved an accuracy of 58.0., which is more than 6\% lower than the original scores of 64.4. Second, we find that few-shot methods outperform zero-shot methods, e.g., LLama 3.1 70B improves from 51.1 to 58.0. Third, we make the interesting observation that the performance is much lower on the human-generated charts (Human (n=390)) than on the automatically-generated charts (Augmented (n=723)). The performance is lower on the original charts and on the LLM-generated charts. Intuitively, this is because human-generated charts generally require much more complex visualization strategies, which makes it more difficult to extract relevant information from VQA models. Hence, the automatically generated charts give a much better performance estimate (chart quality). This can be seen by the little or no differences between the original charts and generated charts for the Human column. As the VQA models improve, human-generated charts will be an important testbed, but because VQA models do not perform well on these examples, automatically generated charts should be the focus. Moreover, there is still a huge gap between the original chart accuracy (80.1) and the best LLM charts (69.9). Fourth, we find that the data regeneration methods only provide a limited view on model performance. With Similarity Score (Sim. Score), all methods perform similarly, with little room for improvement. Moreover, the Exact Match score does not rank models correctly (the best model is ChatGPT Few-Shot (this finding is validated in the error analysis) because of its sensitivity to order and scale.

In Table~\ref{tab:plotqa}, we report the benchmark results on the PlotQA dataset. We make similar findings as the ChartQA dataset, e.g., Few-shot methods outperform Zero-shot. But, on this dataset, performance is similar for both Llama 3.1 70B and GPT3.5. This dataset is synthetic, so overall performance is high for the original charts, making it a good benchmark. Moreover, it has questions subcategorized. For instance, we find a substantial drop in the data-retrieval questions on the LLM-generated charts. We also find that the data extraction results are all over the place on this dataset, thus not providing a good evaluation metric (e.g., exact match is worse on the original charts).

\vspace{2mm} \noindent \textbf{Error Analysis.} In Table~\ref{tab:errors}, we report the results of the manual error analysis of our study. In this study, we focus on the ChartQA dataset errors. We manually reviewed errors made by one model but not the other. These errors were categorized into nine error categories. For example, the VQA Model Error category represents errors where the figures look relatively correct, but the model still did not generate a correct output for some reason. We make two major findings. First, VQA errors are really only a small proportion (based on percentage) of the total errors, and the proportion is relatively similar for both GPT3.5 and Llama errors. For all other errors, we could hypothesize visualization-related issues that caused the errors. Date errors generally result in weird date displays, e.g., 2024.00. Second, the number of errors made by GPT3.5 is larger than Llama, even after accounting for VQA Model Errors. This suggests that the finding that Llaama 70B outperforms GPT3.5 on the ChartQA dataset is a robust and correct finding. \textbf{See the Appendix} for examples of the charts and error types.

\vspace{2mm} \noindent \textbf{Human Study.} In Table~\ref{tab:human}, we report the results of our human study. Specifically, two participants answered survey questions about chart quality (see Methodology section for details). The participants looked at a total of 200 charts each, 100 Llama Few-shot charts and 100 GPT3.5 few-shot charts. The correlation between both participants for the average ratings for both GPT3.5 and LLama on each question was .793 and .853, respectively. As previously mentioned, we categorized these questions into the question types used in the PlotQA dataset. Here we find that the reasoning questions, e.g., ``The LLM-generated chart is easy to read and understand'' perform worse than data retrieval and structural understanding of the charts. These findings match the findings from the PlotQA dataset (e.g., Arithmetic or Compound Questions vs. Data-Retrieval, Min-Max and Structural questions). It is important to note that the participants did not review the original charts, yet the results show the power of the VQA evaluation strategy. We also make the general finding when talking to the participants that the chart design quality leaves a lot to be improved, but is not really explored with this evaluation framework. Future work will be needed to understand and evaluate design quality (e.g., how well the chart looks, which can be impacted by font type and colors).

\section{Conclusion}

We introduced a framework that leverages Visual Question Answering (VQA) models to automate the evaluation of data visualizations generated by Large Language Models (LLMs). This approach addresses the limitations of traditional methods by providing a scalable assessment of both data fidelity and communicative effectiveness. Experiments on the ChartQA and PlotQA datasets revealed that while current LLMs like GPT-3.5 Turbo and Llama 3.1 70B-Instruct do not yet match the accuracy of original non-LLM-generated charts, few-shot prompting significantly enhances their performance. Our error analysis and human study confirmed that VQA models effectively reflect chart quality by capturing visualization issues inherent in LLM-generated charts. This work enables efficient benchmarking and continuous improvement of LLM-driven data visualization systems. There are two major future research directions. First, we can explore how the VQA model performance impacts chart quality estimates. Better aligning and designing the questions that are easy for VQA models from the bottom can make the evaluation more robust. Second, design (e.g., visual quality) is important, and incorporating a chain-of-thought (CoT) structure to improve reasoning, yet completely unexplored here. 

\section*{Limitations}
While our proposed framework provides a scalable method for evaluating LLM-generated data visualizations using VQA models, it has several limitations that warrant discussion.

First, the reliability of our evaluation is inherently tied to the accuracy of the VQA models employed. Current VQA models may not fully capture all the nuances of chart interpretation, especially when dealing with complex visual elements or unconventional chart types. This dependency means that any limitations in VQA accuracy could directly impact the validity of our assessment of the LLM-generated charts.

Second, due to computational and resource constraints, our experiments utilized subsets of the ChartQA and PlotQA datasets. Although these subsets provided valuable insights, they may not fully represent the diversity and complexity of real-world data visualizations. Expanding the scope of our evaluation to include more extensive portions of these datasets—or additional datasets altogether—could enhance the robustness of our findings and provide a more comprehensive understanding of LLM performance in generating accurate and effective visualizations.

Third, the questions used in our evaluation were derived from existing datasets and may not be perfectly aligned with the specific visual aspects of the charts generated by the LLMs. This misalignment could lead to an incomplete or skewed assessment of chart quality. Future work should focus on developing more targeted questions that are specifically designed to evaluate particular visual attributes or design elements of the generated charts. By aligning question types more closely with the visual features being assessed, we can obtain more precise estimates of chart quality and better identify areas where LLMs excel or require improvement.

Addressing these limitations will be crucial for refining our evaluation framework and enhancing its applicability. Improvements in VQA model accuracy, the use of larger and more diverse datasets, and the development of tailored evaluation questions will collectively contribute to a more robust and insightful assessment of LLM-generated data visualizations.

\bibliography{custom}

\appendix

\section{Appendix}
\label{sec:appendix}
The following examples illustrate where the charts generated by ChatGPT were not correctly evaluated by the VQA model while the version generated by Llama were.  The major categories from Table 6 are depicted. 

\vspace{2mm} \noindent \textbf{Chart Visualization Errors: Category Ambiguous}

\raggedbottom 

Question: What percentage of COVID-19 patients died after contracting the virus?

\begin{figure}[H] 
\centering
\begin{subfigure}{0.5\textwidth}
\centering
\includegraphics[width=\linewidth]{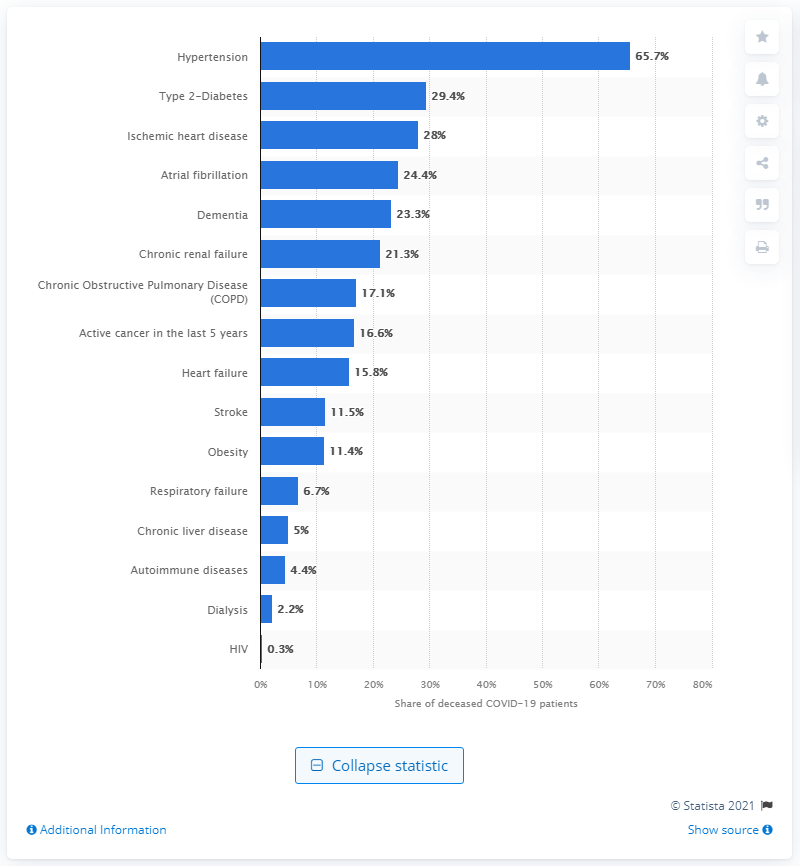}
\caption{Ground Truth}
\label{fig:appendix_groundtruth}
\end{subfigure}
\hfill
\begin{subfigure}{0.5\textwidth}
\centering
\includegraphics[width=\linewidth]{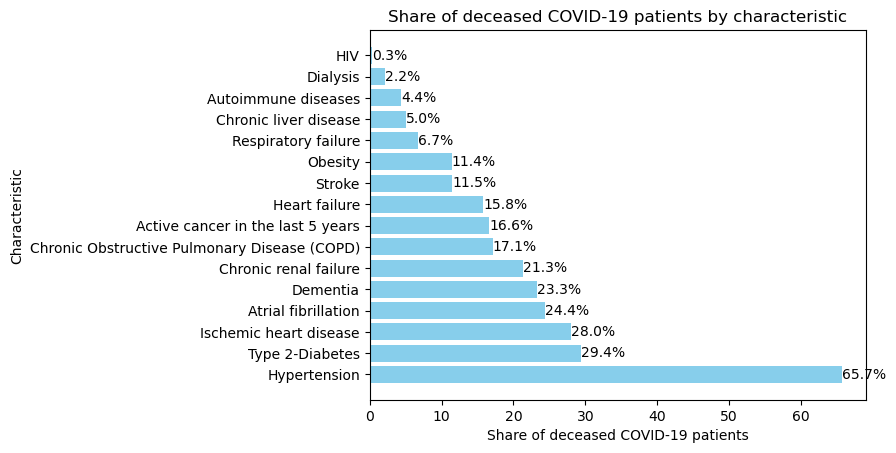}
\caption{ChatGPT}
\label{fig:appendix_chatgpt}
\end{subfigure}
\hfill
\begin{subfigure}{0.5\textwidth}
\centering
\includegraphics[width=\linewidth]{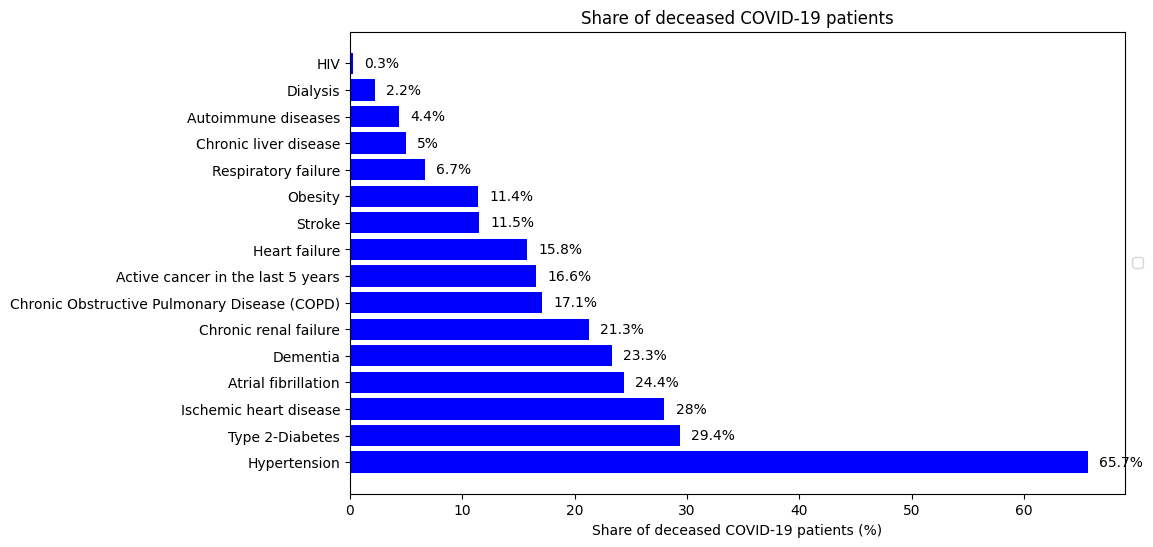}
\caption{Llama}
\label{fig:appendix_llama}
\end{subfigure}
\caption{Chart Visualization Errors: Category Ambiguous}
\label{fig:chart_errors}
\end{figure}

Because the question did not specifically ask for which condition led to death, the VQA model had to choose which percentage to report, as shown in Figure~\ref{fig:chart_errors}.

\vspace{2mm} \noindent \textbf{Chart Visualization Errors: Colors Not Matching}





\raggedbottom

\textbf{Question:} What percentage is represented by the navy blue bar?

\begin{figure}[H]
\centering
\begin{subfigure}{0.5\textwidth}
\centering
\includegraphics[width=\linewidth]{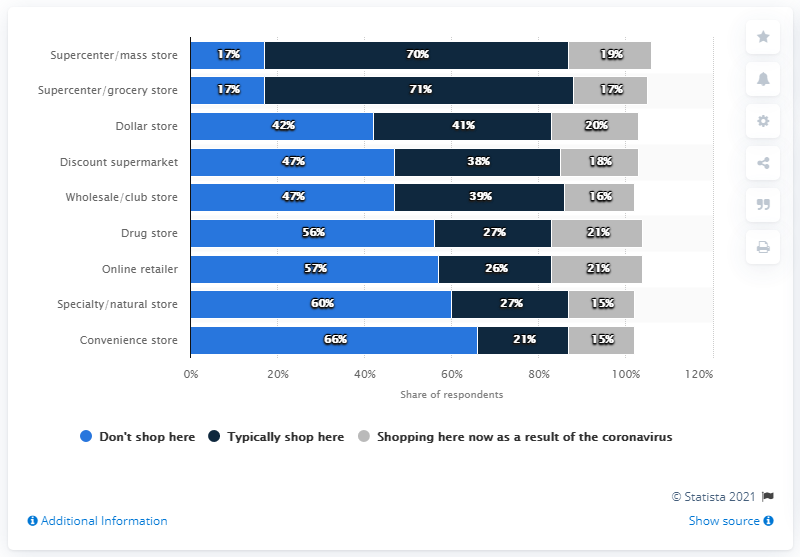}
\caption{Ground Truth}
\label{fig:appendix_groundtruth}
\end{subfigure}
\hfill
\begin{subfigure}{0.5\textwidth}
\centering
\includegraphics[width=\linewidth]{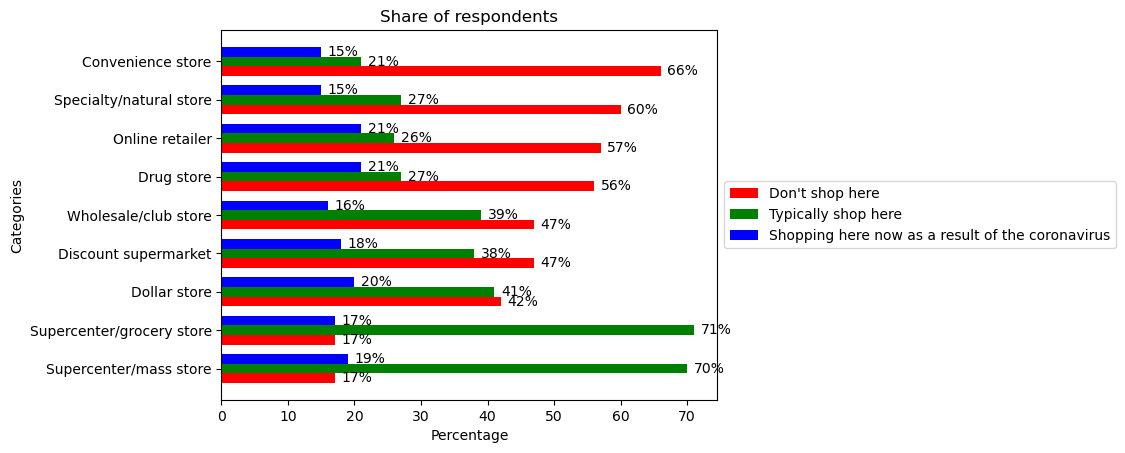}
\caption{ChatGPT}
\label{fig:appendix_chatgpt}
\end{subfigure}
\hfill
\begin{subfigure}{0.5\textwidth}
\centering
\includegraphics[width=\linewidth]{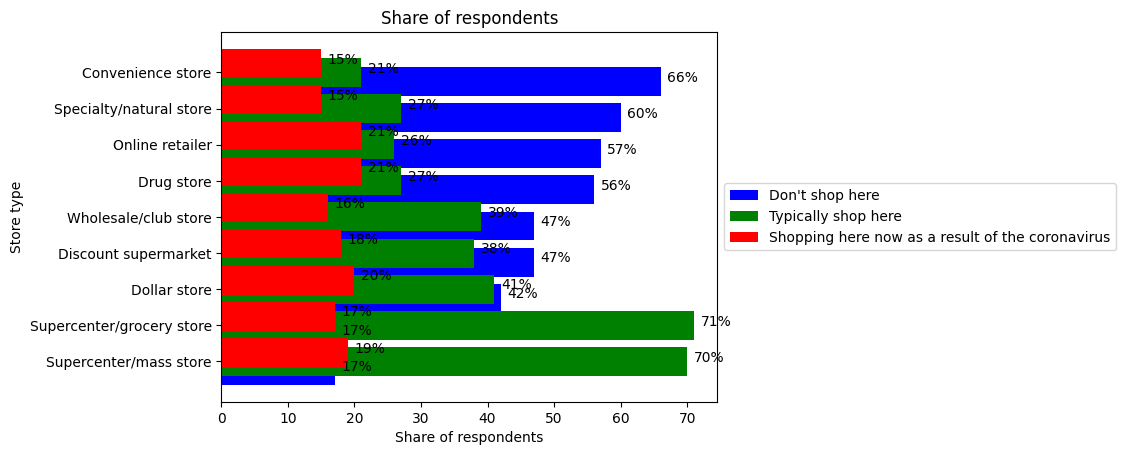}
\caption{Llama}
\label{fig:appendix_llama}
\end{subfigure}
\caption{Chart Visualization Errors: Colors Not Matching}
\label{fig:chart_errors_missmatching}
\end{figure}

The colors used in the ground truth charts were not provided in the instructions to the LLMs, so the generated-charts did not necessarily match the original, as shown in Figure~\ref{fig:chart_errors_missmatching}.

\vspace{2mm} \noindent \textbf{Chart Visualization Errors: Dates Errors}

\raggedbottom

\textbf{Question:} What was the Polish gender equality index score between 2005 and 2019?

\begin{figure}[H]
\centering
\begin{subfigure}{0.5\textwidth}
\centering
\includegraphics[width=\linewidth]{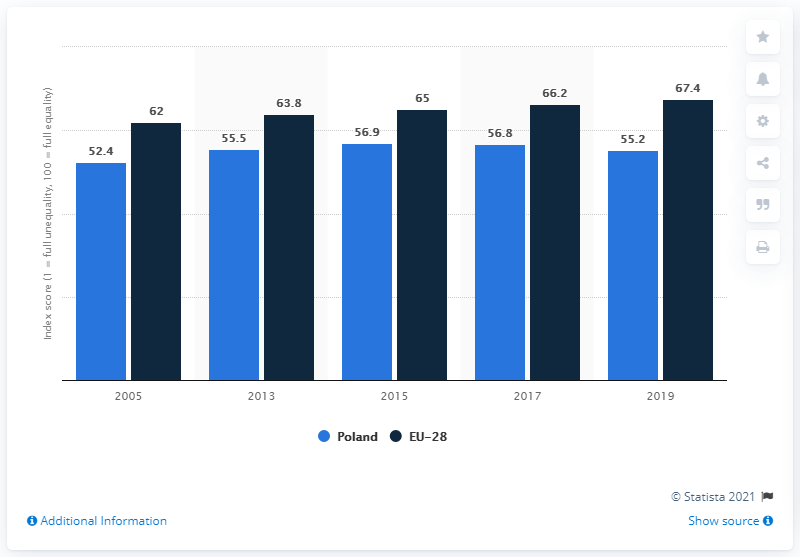}
\caption{Ground Truth}
\label{fig:appendix} 
\end{subfigure}

\begin{subfigure}{0.5\textwidth}
\centering
\includegraphics[width=\linewidth]{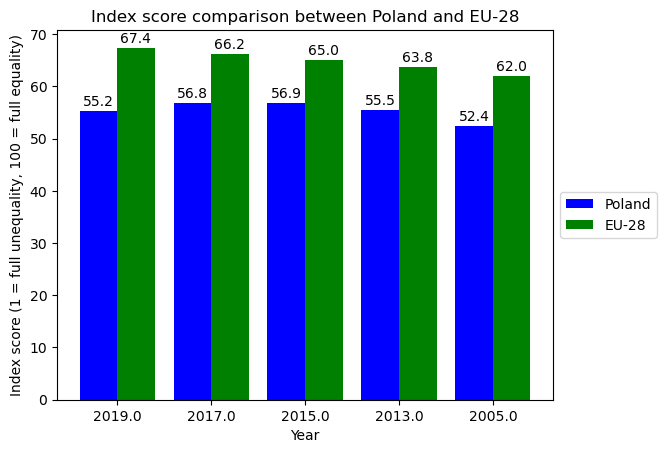}
\caption{ChatGPT}
\label{fig:appendix}
\end{subfigure}

\begin{subfigure}{0.5\textwidth}
\centering
\includegraphics[width=\linewidth]{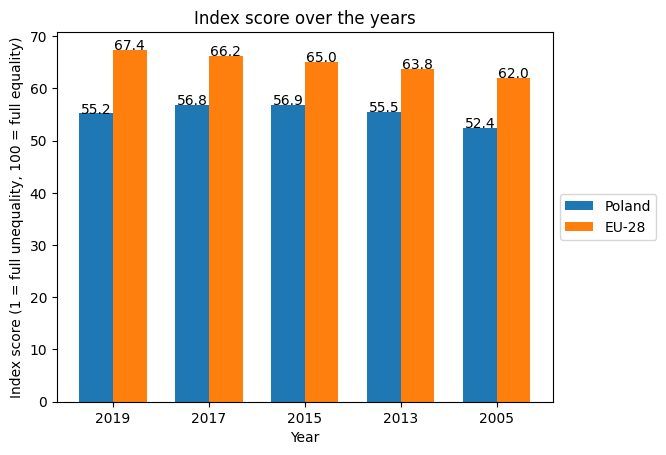}
\caption{Llama}
\label{fig:appendix}
\end{subfigure}
\caption{Chart Visualization Errors:  Dates Errors}
\label{fig:chart_vierrors}
\end{figure}

The date labels on the X-axis are not correct on the ChatGPT-generated chart, leading to incorrect answers from the VQA model, as shown in Figure~\ref{fig:chart_vierrors}.

\vspace{2mm} \noindent \textbf{Chart Visualization Errors: Labels Overlapping }

\raggedbottom

\textbf{Question:}  How many daily active users did Douyin have in comparison to the period prior to the epidemic?

\begin{figure}[H]
\centering
\begin{subfigure}{0.5\textwidth}
\centering
\includegraphics[width=\linewidth]{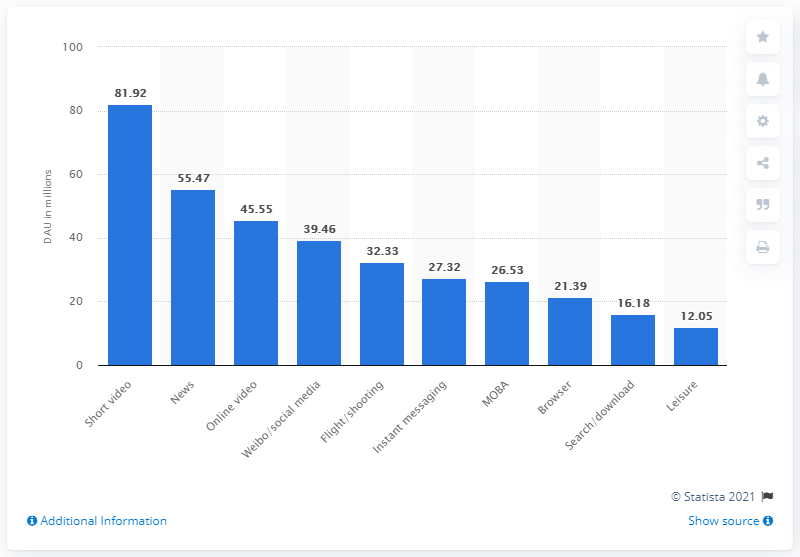}
\caption{Ground Truth}
\label{fig:appendix} 
\end{subfigure}

\begin{subfigure}{0.5\textwidth}
\centering
\includegraphics[width=\linewidth]{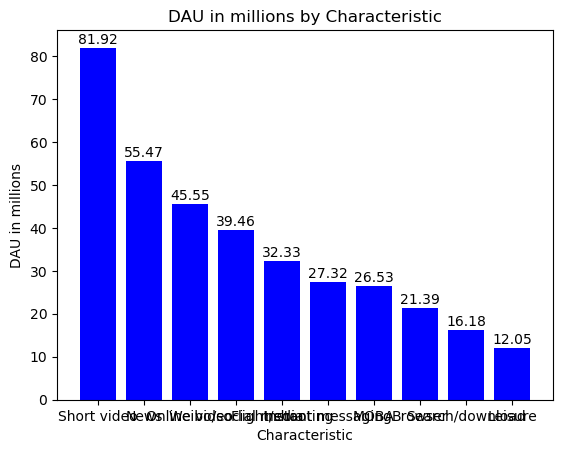}
\caption{ChatGPT}
\label{fig:appendix}
\end{subfigure}

\begin{subfigure}{0.5\textwidth}
\centering
\includegraphics[width=\linewidth]{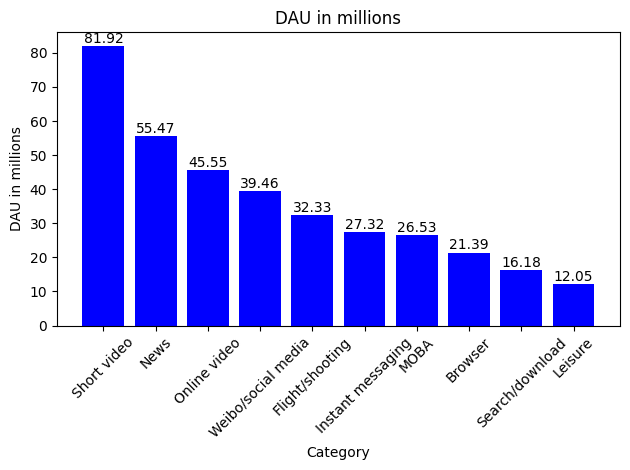}
\caption{Llama}
\label{fig:appendix}
\end{subfigure}
\caption{Chart Visualization Errors:  Labels Overlapping}
\label{fig:chart_errors_ol}
\end{figure}

The category labels on the X-axis are not correct on the ChatGPT-generated chart, leading to incorrect answers from the VQA model, as shown in Figure~\ref{fig:chart_errors_ol}.

\vspace{2mm} \noindent \textbf{Chart Visualization Errors: Wrong Type of Bars}

\raggedbottom

\textbf{Question:} How many bars (combined) in the chart?

\begin{figure}[H]
\centering
\begin{subfigure}{0.5\textwidth}
\centering
\includegraphics[width=\linewidth]{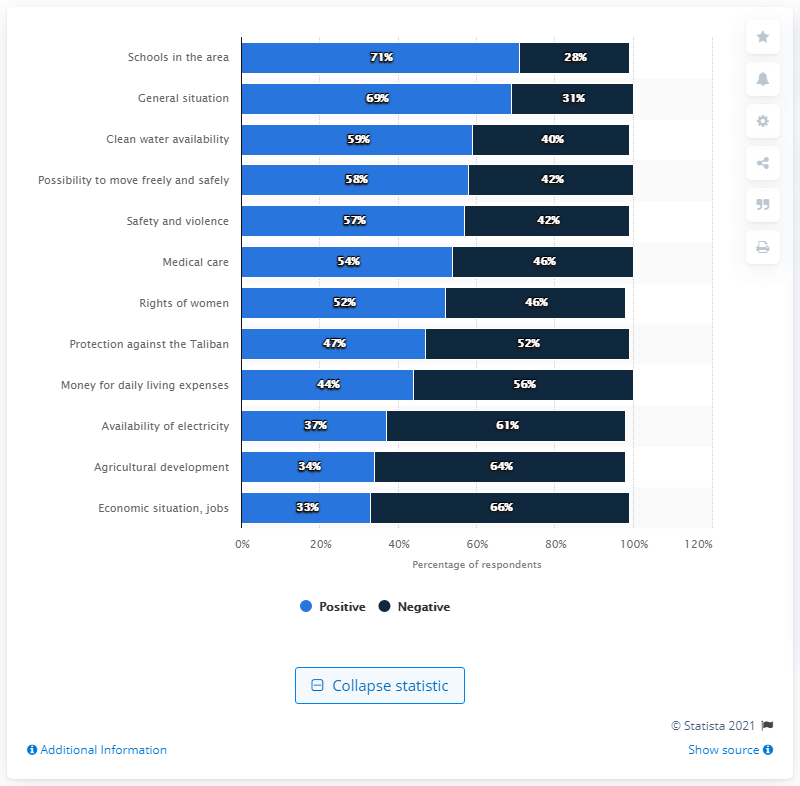}
\caption{Ground Truth}
\label{fig:appendix} 
\end{subfigure}

\begin{subfigure}{0.5\textwidth}
\centering
\includegraphics[width=\linewidth]{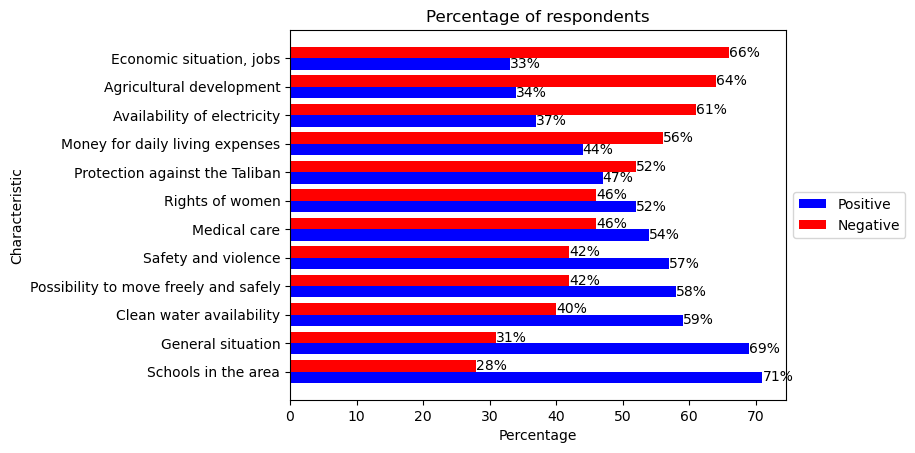}
\caption{ChatGPT}
\label{fig:appendix}
\end{subfigure}

\begin{subfigure}{0.5\textwidth}
\centering
\includegraphics[width=\linewidth]{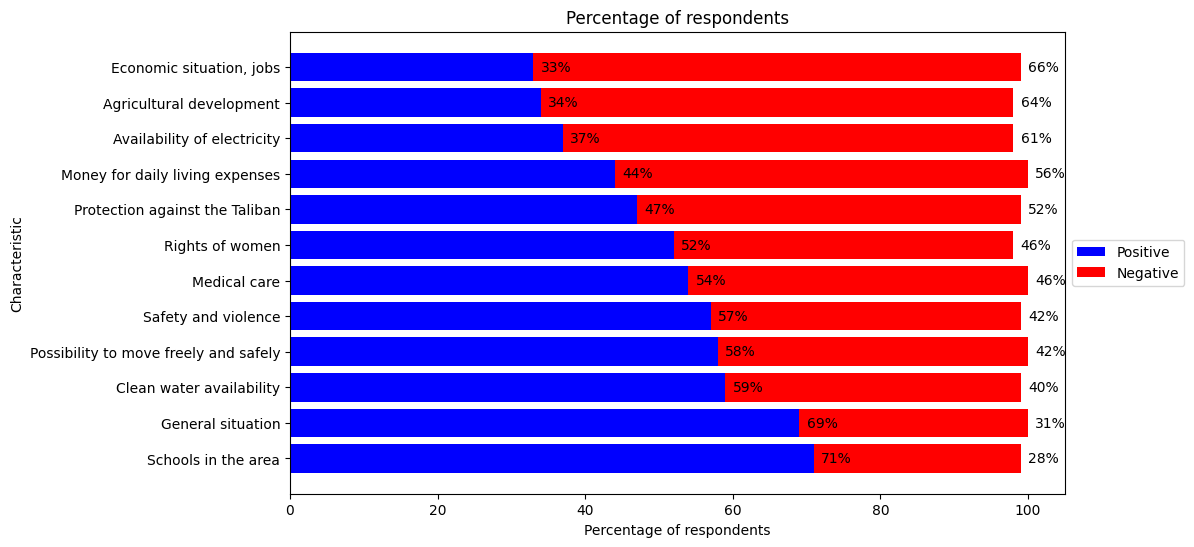}
\caption{Llama}
\label{fig:appendix}
\end{subfigure}
\caption{Chart Visualization Errors:  Wrong Type of Bars}
\label{fig:chart_errors_type}
\end{figure}

The ChatGPT-generated chart does not utilize stacked bars, leading to incorrect answers from the VQA model, as shown in Figure~\ref{fig:chart_errors_type}.

\end{document}